\documentclass[10pt,twocolumn,letterpaper]{article}

\usepackage{iccv}
\usepackage{times}
\usepackage{epsfig}
\usepackage{graphicx}
\usepackage{amsmath}
\usepackage{amssymb}
\usepackage{subfigure}
\usepackage{csquotes}

\usepackage{multirow}
\usepackage{graphicx}
\usepackage{booktabs}
\usepackage{bbm}

\usepackage{color, colortbl}
\newcommand{\norm}[1]{\left\|#1\right\|}
\newcommand{\inner}[2]{#1\cdot #2} %inner product

\newcommand*{\twoelementtable}[3][l]%
{%  
\renewcommand{\arraystretch}{0.8}%
\begin{tabular}[t]{@{}#1@{}}%
#2\tabularnewline
#3%
\end{tabular}%
}
\usepackage{array}
\usepackage{tabularx}

\iccvfinalcopy % *** Uncomment this line for the final submission

\usepackage[breaklinks=true,bookmarks=false]{hyperref}

\def\iccvPaperID{2104} % *** Enter the ICCV Paper ID here

\ificcvfinal\fi

\begin{document}

\title{Domain Adaptive Video Segmentation via Temporal Consistency Regularization}

\author{Dayan Guan, Jiaxing Huang, Aoran Xiao, Shijian Lu\thanks{Corresponding author.} \\ 
School of Computer Science Engineering, Nanyang Technological University\\
{\tt\small \{Dayan.Guan, Jiaxing.Huang, Aoran.Xiao, Shijian.Lu\}@ntu.edu.sg}
}

\maketitle
\ificcvfinal\thispagestyle{empty}\fi

\begin{abstract}
Video semantic segmentation is an essential task for the analysis and understanding of videos. Recent efforts largely focus on supervised video segmentation by learning from fully annotated data, but the learnt models often experience clear performance drop while applied to videos of a different domain. This paper presents DA-VSN, a domain adaptive video segmentation network that addresses domain gaps in videos by temporal consistency regularization (TCR) for consecutive frames of target-domain videos. DA-VSN consists of two novel and complementary designs. The first is cross-domain TCR that guides the prediction of target frames to have similar temporal consistency as that of source frames (learnt from annotated source data) via adversarial learning. The second is intra-domain TCR that guides unconfident predictions of target frames to have similar temporal consistency as confident predictions of target frames. Extensive experiments demonstrate the superiority of our proposed domain adaptive video segmentation network which outperforms multiple baselines consistently by large margins.
\end{abstract}

\section{Introduction}

\begin{figure}
\centering
\includegraphics[width=.98\linewidth]{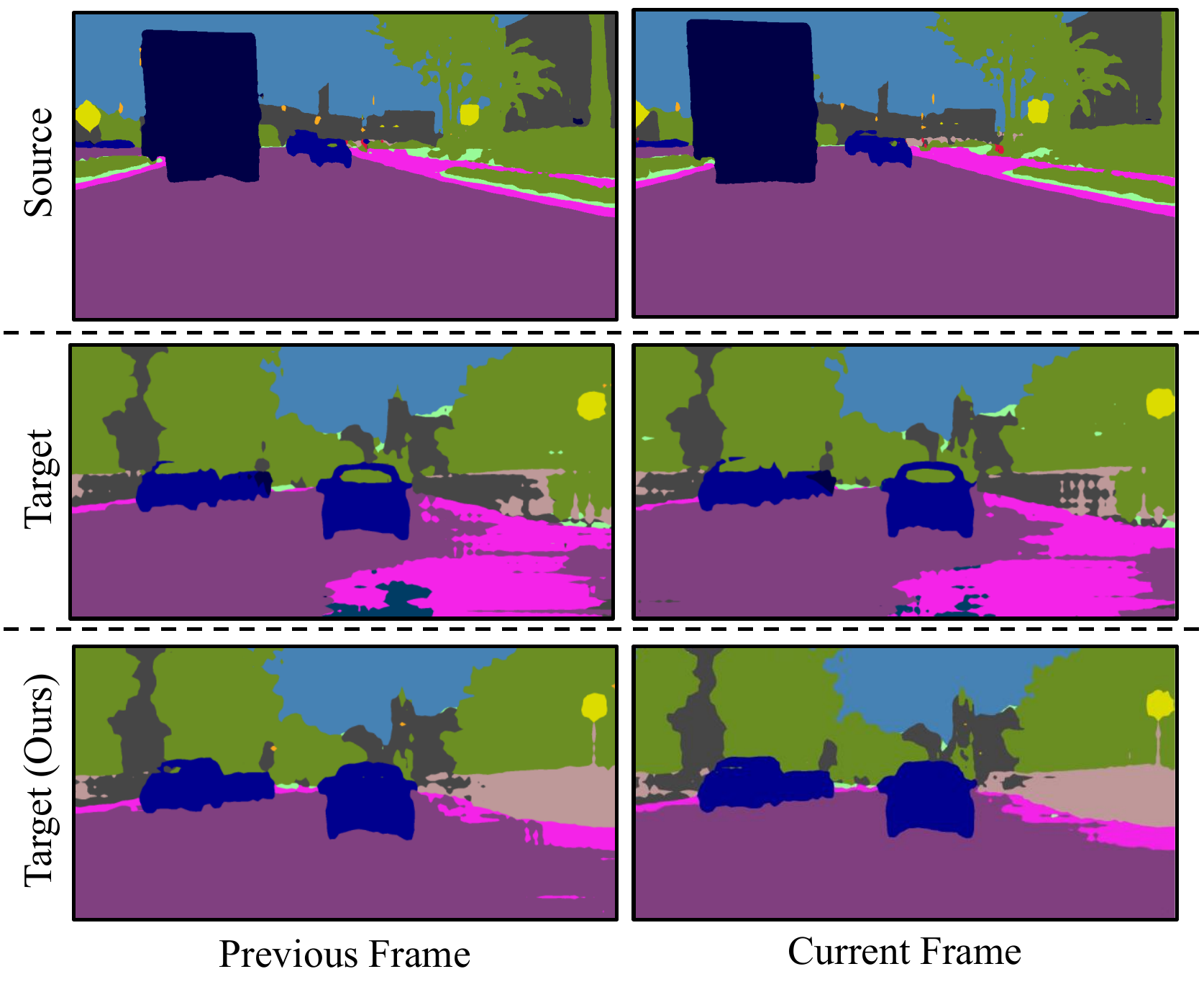}
\caption{Temporal consistency helps in domain adaptive video segmentation: A video segmentation model trained in a \textit{Source} domain often experiences clear performance drop while applied to videos of a \textit{Target} domain. We employ temporal consistency, the inherent and universal nature of videos, as a constraint to regularize inter-domain and intra-domain adaptation for optimal video segmentation in target domain as in \textit{Target (ours)}.}
\label{fig:intro}
\end{figure}

Video semantic segmentation aims to assign pixel-wise semantic labels to video frames, and it has been attracting increasing attention as one essential task in video analysis and understanding~\cite{floros2012joint,miksik2013efficient,couprie2013causal,liu2015multiclass,patraucean2015spatio}. With the advance of deep neural networks (DNNs), several studies have been conducted in recent years with very impressive video segmentation performance~\cite{shelhamer2016clockwork,kundu2016feature,gadde2017semantic,huang2018efficient,jain2019accel,kim2020video,hu2020temporally,liu2020efficient}. However, most existing works require large amounts of densely annotated training videos which entail a prohibitively expensive and time-consuming annotation process~\cite{brostow2008segmentation,cordts2016cityscapes}. One approach to alleviate the data annotation constraint is to resort to self-annotated synthetic videos that are collected with computer-generated virtual scenes~\cite{richter2017playing,hernandez2017slanted}, but models trained with the synthesized data often experience clear performance drops while applied to videos of natural scenes largely due to the \textit{domain shift} as illustrated in Fig.~\ref{fig:intro}.

Domain adaptive video segmentation is largely neglected in the literature despite its great values in both research and practical applications. It could be addressed from two approaches by leveraging existing research. The first approach is domain adaptive image segmentation ~\cite{zou2018unsupervised,vu2019advent,zou2019confidence,pan2020unsupervised,yang2020fda} which could treat each video frame independently to achieve domain adaptive video segmentation. However, domain adaptive image segmentation does not consider temporal information in videos which is very important in video semantic segmentation. The second approach is semi-supervised video segmentation ~\cite{nilsson2018semantic,zhu2019improving,chen2020naive} that exploits sparsely annotated video frames for segmenting unannotated frames of the same video. However, semi-supervised video segmentation was designed for consecutive video frames of the same domain and does not work well in domain adaptive video segmentation which usually involves clear domain shifts and un-consecutive video frames of different sources.

In this work, we design a domain adaptive video segmentation network (DA-VSN) that introduces temporal consistency regularization (TCR) to bridge the gaps between videos of different domains. The design is based on the observation that video segmentation model trained in a source domain tends to produce temporally consistent predictions over source-domain data but temporally inconsistent predictions over target-domain data (due to domain shifts) as illustrated in Fig.~\ref{fig:intro}. We designed two complementary regularization modules in DA-VSN, namely, cross-domain TCR (C-TCR) and intra-domain TCR (I-TCR). C-TCR employs adversarial learning to minimize the discrepancy of temporal consistency between source and target domains. Specifically, it guides target-domain predictions to have similar temporal consistency of source-domain predictions which usually has decent quality by learning from fully-annotated source-domain data. I-TCR instead works from a different perspective by guiding unconfident target-domain predictions to have similar temporal consistency as confident target-domain predictions. In I-TCR, we leverage entropy to measure the prediction confidence which works effectively across multiple datasets.

The contributions of this work can be summarized in three major aspects. \textit{First}, we proposed a new framework that introduces temporal consistency regularization (TCR) to address domain shifts in domain adaptive video segmentation. To the best of our knowledge, this is the first work that tackles the challenge of unsupervised domain adaptation in video semantic segmentation. \textit{Second}, we designed inter-domain TCR and intra-domain TCR that improve domain adaptive video segmentation greatly by minimizing the discrepancy of temporal consistency across different domains and different video frames in target domain, respectively. \textit{Third}, extensive experiments over two challenging synthetic-to-real benchmarks (VIPER~\cite{richter2017playing}~$\rightarrow$~Cityscapes-Seq~\cite{cordts2016cityscapes} and SYNTHIA-Seq~\cite{ros2016synthia}~$\rightarrow$~Cityscapes-Seq) show that the proposed DA-VSN achieves superior domain adaptive video segmentation as compared with multiple baselines.

\section{Related Works}

\subsection{Video Semantic Segmentation}
Video semantic segmentation aims to predict pixel-level semantics for each video frame. Most existing works exploit inter-frame temporal relations for robust and accurate segmentation~\cite{hur2016joint,zhu2017deep,gadde2017semantic,liu2017surveillance,xu2018dynamic,li2018low,jain2019accel,hu2020temporally,liu2020efficient}. For example, \cite{zhu2017deep,gadde2017semantic} employs optical flow~\cite{dosovitskiy2015flownet} to warp feature maps between frames. \cite{li2018low} leverages inter-frame feature propagation for efficient video segmentation with low latency. \cite{jain2019accel} presents an adaptive fusion policy for effective integration of predictions from different frames. \cite{hu2020temporally} distributes several sub-networks over sequential frames and recomposes the extracted features via attention propagation. \cite{liu2020efficient} presents a compact network that distills temporal consistency knowledge for per-frame inference.

In addition, semi-supervised video segmentation has been investigated which exploits sparsely annotated video frames for segmenting unannotated frames of the same videos. Two typical approaches have been studied. The first approach is based on label propagation that warps labels of sparsely-annotated frames to generate pseudo labels for unannotated frames via patch matching~\cite{badrinarayanan2010label,budvytis2017large}, motion cues~\cite{tokmakov2016weakly,zhu2019improving} or optical flow~\cite{mustikovela2016can,zhu2017deep,nilsson2018semantic,ding2020every}. The other approach is based on self-training that generates pseudo labels through a distillation across multiple augmentations~\cite{chen2020naive}. 

Both supervised and semi-supervised video segmentation work on frames of the same video or same domain that have little domain gaps. In addition, they both require a certain amount of pixel-level annotated video frames that are prohibitively expensive and time-consuming to collect. Our proposed domain adaptive video segmentation exploits off-the-shelf video annotations from a \textit{source domain} for the segmentation of videos of a different \textit{target domain} without requiring any annotations of target-domain videos.

\subsection{Domain Adaptive Video Classification} Domain adaptive video classification has been explored to investigate domain discrepancy in action classification problem. One category of works focuses on the specific action recognition task that aims to classify a video clip into a particular category of human actions via temporal alignment~\cite{chen2019temporal}, temporal attention~\cite{pan2020adversarial, choi2020shuffle}, or self-supervised video representation learning~\cite{munro2020multi, choi2020shuffle}. Another category of works focus on action segmentation that simultaneously segments a video in time and classifies each segmented video clip with an action class via temporal alignment~\cite{chen2020action2} or self-supervised video representation learning~\cite{chen2020action}.

This work focuses on a new problem of domain adaptive semantic segmentation of videos, a new and much more challenging domain adaptation task as compared with domain adaptive video classification. Note that existing domain adaptive video classification methods do not work for the semantic segmentation task as they cannot generate pixel-level dense predictions for each frame in videos. 

\begin{figure*}[!ht]
\centering
\includegraphics[width=.9\linewidth]{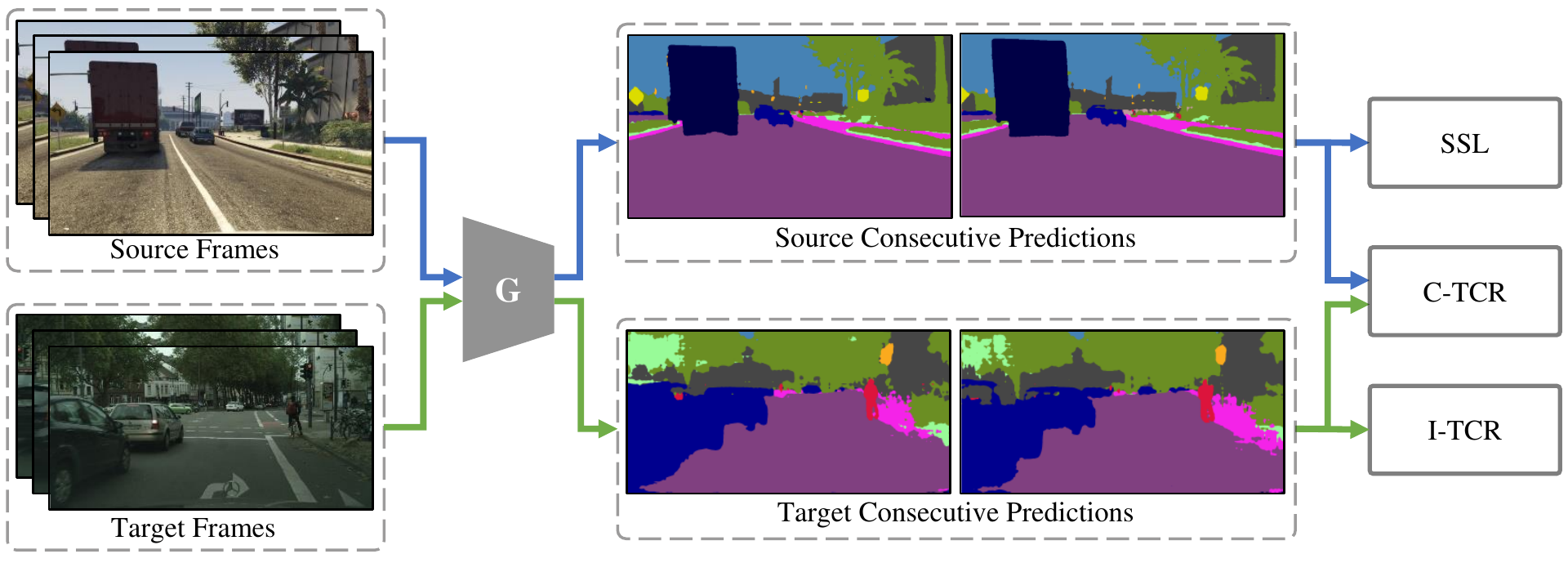}
\caption{The framework of the proposed domain adaptive video segmentation network (DA-VSN): DA-VSN introduces temporal consistency regularization (TCR) to minimize the divergence between source and target domains. It consists of a video semantic segmentation model \textit{G} that generates segmentation predictions, a source-domain supervised learning module (\textit{SSL}) that learns knowledge from source domain, a cross-domain TCR component (\textit{C-TCR}) that guides target predictions to have similar temporal consistency as source predictions, and an intra-domain TCR component (\textit{I-TCR}) that guides unconfident target predictions to have similar temporal consistency as confident target predictions.}
\label{fig:DA-VSN}
\vspace{-2pt}
\end{figure*} 

\subsection{Domain Adaptive Image Segmentation}
Domain adaptive image segmentation has been widely investigated to address the image annotation challenge and domain shift issues~\cite{chen2019domain,yang2020fda}. Most existing methods take two typical approaches, namely, adversarial learning based~\cite{hoffman2018cycada,tsai2018learning,vu2019advent,tsai2019domain,luo2019significance,huang2020contextual,pan2020unsupervised,guan2021scale} and self training based~\cite{zou2018unsupervised,zou2019confidence,li2019bidirectional,lian2019constructing,yang2020fda,kim2020learning,zheng2021rectifying,mei2020instance}. The adversarial learning based methods perform domain alignment by adopting a discriminator that strives to differentiate the segmentation in the space of inputs~\cite{hoffman2018cycada,zhang2018fully,li2019bidirectional,choi2019self,kim2020learning}, features~\cite{tzeng2017adversarial,hoffman2018cycada,chen2018road,zhang2018fully,luo2019significance} or outputs~\cite{tsai2018learning,vu2019advent,luo2019taking,tsai2019domain,huang2020contextual,lv2020cross,wang2020differential,pan2020unsupervised,guan2021scale}. The self-training based methods exploit self-training to predict pseudo labels for target-domain data and then exploit the predicted pseudo labels to fine-tune the segmentation model iteratively.

Though a number of domain adaptive image segmentation techniques have been reported in recent years, they do not consider temporal information which is critically important in video segmentation. We introduce temporal consistency of videos as a constraint and exploit it to regularize the learning in domain adaptive video segmentation. 

\section{Method}

\subsection{Problem Definition}
Given source-domain video frames $X^{\mathbb{S}}$ with the corresponding labels $Y^{\mathbb{S}}$ and target-domain video frames $X^{\mathbb{T}}$ without labels, the goal of domain adaptive video segmentation is to learn a model $G$ that can produce accurate predictions $P^{\mathbb{T}}$ in target domain. According to the domain adaptation theory in~\cite{ben2010theory}, the target error in domain adaptation is bounded by three terms including a shared error of the ideal joint hypothesis on the source and target domains, an empirical source-domain error, and a divergence measure between source and target domains.

This work focuses on the third term and presents a domain adaptive video semantic segmentation network (DA-VSN) for minimizing the divergence between source and target domains. We design a novel temporal consistency regularization (TCR) technique for consecutive frames in target domain, which consists of two complementary components including a cross-domain TCR (C-TCR) component and an intra-domain TCR (I-TCR) component as illustrated in Fig.~\ref{fig:DA-VSN}. C-TCR targets cross-domain alignment by encouraging target predictions to have similar temporal consistency as source predictions (accurate via supervised learning), while I-TCR aims for intra-domain adaptation by forcing unconfident predictions to have similar temporal consistency as confident predictions in target domain, more details to be described in the ensuing two subsections.

Note the shared error in the first term (the difference in labeling functions across domains) is usually small as proven in~\cite{ben2010theory}. The empirical source-domain error in the second term actually comes from the supervised learning in the source domain. For domain adaptive video segmentation, we directly adopt video semantic segmentation loss $\mathcal{L}_{ssl} (G)$~\cite{zhu2017deep, jain2019accel,hu2020temporally,liu2020efficient} as the source-domain supervised learning loss.

\begin{figure}[!ht]
\centering
\includegraphics[width=.99\linewidth]{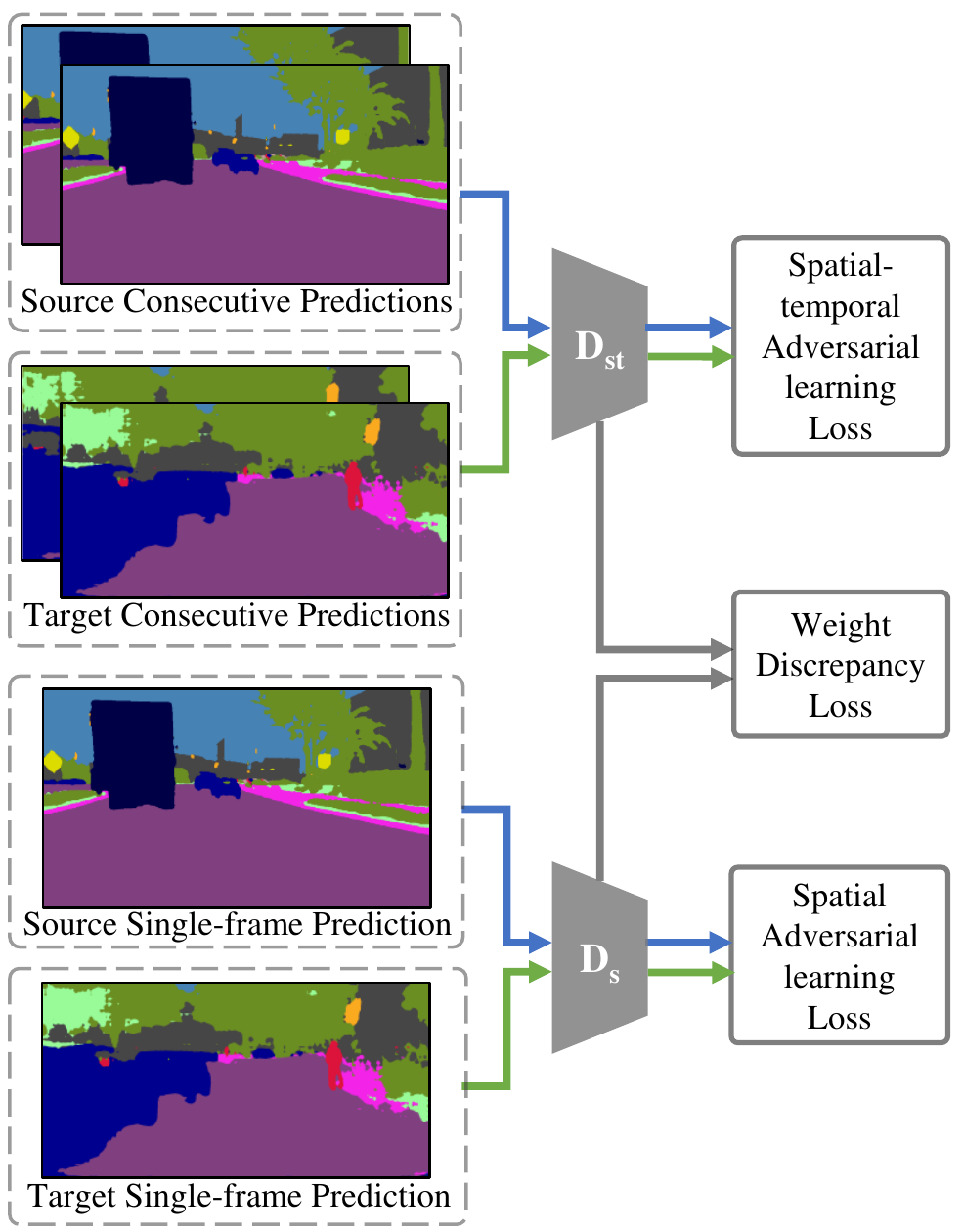}
\caption{The framework of the proposed cross-domain temporal consistency regularization (C-TCR): C-TCR performs temporal alignment to minimize the divergence of temporal consistency between source and target domains. It introduces a spatial-temporal discriminator $D_{st}$ to align consecutive predictions (encoding spatial-temporal information) and a spatial discriminator $D_{s}$ to align single-frame predictions (encoding spatial information). A spatial-temporal adversarial learning loss $\mathcal{L}_{sta}$ and a spatial loss $\mathcal{L}_{sa}$ are introduced to optimize the discriminators and segmentation model. To enhance temporal alignment, we introduce a weight discrepancy loss $\mathcal{L}_{wd}$ to force $D_{st}$ to be independent from $D_{s}$ so that $D_{st}$ can focus more on temporal alignment.}
\label{fig:C-TCR}
\end{figure}

\subsection{Cross-domain Regularization}
Cross-domain temporal consistency regularization (C-TCR) aims to guide target predictions to have similar temporal consistency of source predictions which is determined by minimizing the supervised source loss $\mathcal{L}_{ssl}$ and usually has decent quality. We design a dual-discriminator structure for optimal spatial-temporal alignment of source and target \textit{video-clips} as illustrated in Fig.~\ref{fig:C-TCR}. As Fig.~\ref{fig:C-TCR} shows, one discriminator $D_s$ focuses on spatial alignment of a single video frame of different domains (as in domain adaptive image segmentation) and the other discriminator $D_{st}$ focuses on temporal alignment of consecutive videos frames of different domains. Since $D_{st}$ inevitably involves spatial information, we introduce a divergence loss between $D_s$ and $D_{st}$ to force $D_{st}$ to focus on the alignment in temporal space.

For spatial alignment, the spatial discriminator $D_s$ aligns frame-level predictions $p^{\mathbb{S}}_{k}$ and $p^{\mathbb{T}}_{k}$ and its objective $\mathcal{L}_{sa}$ can be formulated as follows:
\begin{equation}
\begin{split}
\mathcal{L}_{sa} (G,D_{s}) = \log (D_{s}(p^{\mathbb{S}}_{k})) + \log (1 - D_{s}(p^{\mathbb{T}}_{k})).
\end{split}
\label{eq:sa}
\end{equation}

For temporal alignment, we forward the current frame $x^{\mathbb{S}}_{k}$ and the consecutive frame $x^{\mathbb{S}}_{k-1}$ to obtain the current prediction $p^{\mathbb{S}}_{k}$, and simultaneously forward $x^{\mathbb{S}}_{k-1}$ and $x^{\mathbb{S}}_{k-2}$ to obtain its consecutive prediction $p^{\mathbb{S}}_{k-1}$. The two consecutive predictions are stacked as $p^{\mathbb{S}}_{k-1:k}$ which encode spatial-temporal information in source domain. This same process is applied to target domain which produces two consecutive target predictions $p^{\mathbb{T}}_{k-1:k}$ that encode spatial-temporal information in target domain. The spatial-temporal discriminator $D_{st}$ then aligns $p^{\mathbb{S}}_{k-1:k}$ and $p^{\mathbb{T}}_{k-1:k}$ and its objective $\mathcal{L}_{sta}$ can be formulated as follows:
\begin{equation}
\begin{split}
\mathcal{L}_{sta} (G,D_{st}) 
& = \log (D_{sta}(p^{\mathbb{S}}_{k-1:k})) \\
& + \log (1 - D_{sta}(p^{\mathbb{T}}_{k-1:k})). 
\end{split}
\label{eq:sta}
\end{equation}

We enforce the divergence of the weights of $D_{st}$ and $D_{s}$ so that the spatial-temporal discriminator $D_{st}$ can focus more on temporal alignment. The weight divergence of the two discriminators can be reduced by minimizing their cosine similarity as follows:
\begin{equation}
\begin{split}
\mathcal{L}_{wd} (D_{st},D_{s}) = \frac{1}{J} \sum_{j=1}^{J} \frac{\inner{\overrightarrow{w}^{j}_{st}}{\overrightarrow{w}^{j}_{s}}}{\norm{\overrightarrow{w}^{j}_{st}}\norm{\overrightarrow{w}^{j}_{s}}},
\end{split}
\label{eq:wd}
\end{equation}
where $J$ is the number of convolutional layers in each discriminator, $\overrightarrow{w}^{j}_{st}$ and $\overrightarrow{w}^{j}_{s}$ are obtained by flattening the weights of the $j$-th convolutional layer
in the discriminators $D_{st}$ and $D_{s}$, respectively. 

Combining the losses in Eqs~\ref{eq:sa},~\ref{eq:sta},~\ref{eq:wd}, the C-TCR loss $\mathcal{L}_{ctcr}$ can be formulated as follows:
\begin{equation}
\begin{split}
\mathcal{L}_{ctcr} (G,D_{st},D_{s}) 
& = \mathcal{L}_{sta} (G,D_{st}) \\
& + \lambda_{sa} \mathcal{L}_{sa} (G,D_{s}) \\
& + \lambda_{wd} \mathcal{L}_{wd} (D_{st},D_{s}),
\end{split}
\label{eq:ctcr}
\end{equation}
where $\lambda_{sa}$ and $\lambda_{wd}$ are the balancing weights.

\begin{figure*}[!ht]
\centering
\includegraphics[width=.9\linewidth]{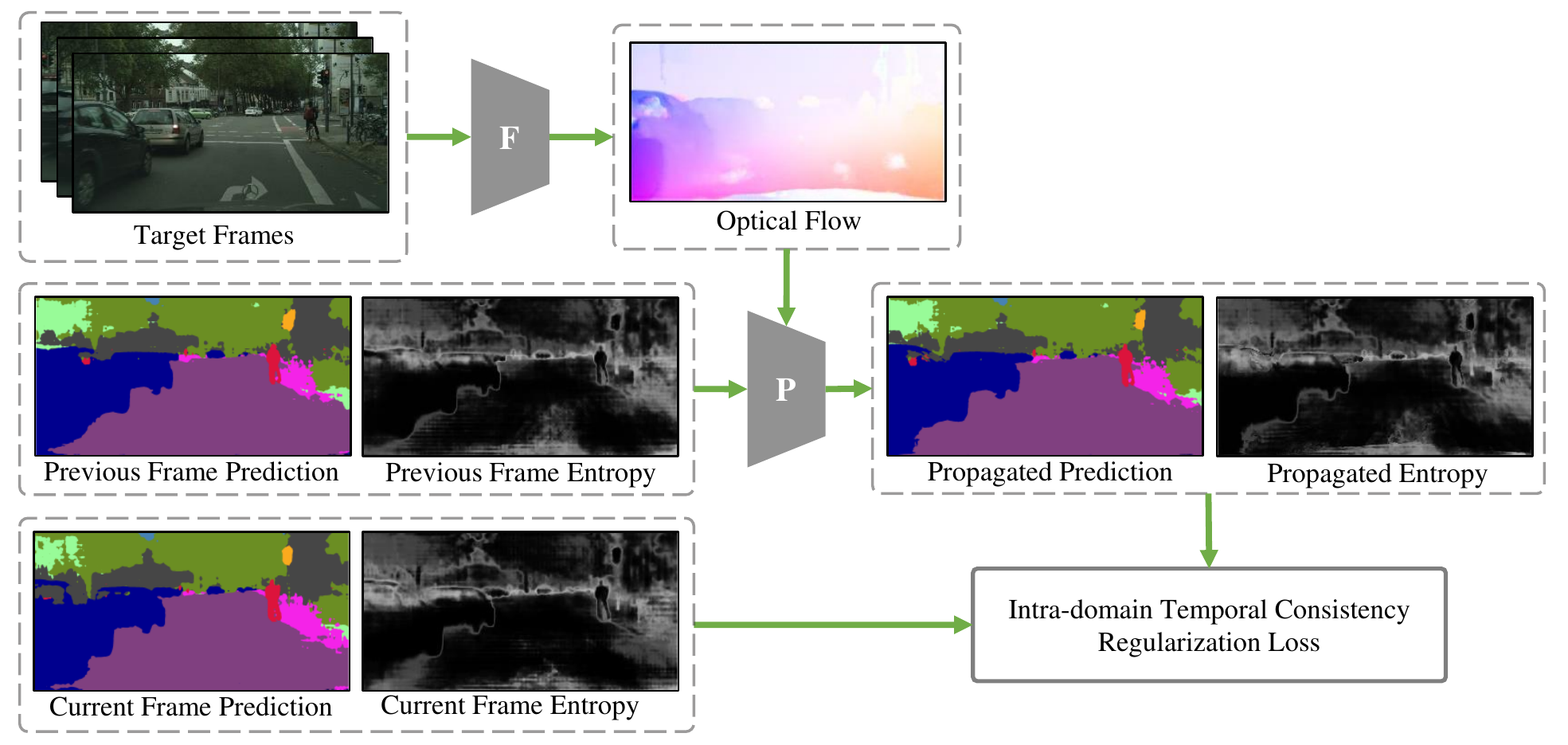}
\caption{The framework of the proposed intra-domain temporal consistency regularization (I-TCR): I-TCR guides unconfident target-domain predictions to have similar temporal consistency as confident predictions, where the prediction confidence is measured by entropy. It consists of a network $F$ to estimate optical flow and a propagation operation $P$ to warp the previous frame prediction and its entropy based on the estimated optical flow. The objective of I-TCR is to force the current frame prediction with high entropy (\ie, low confidence) to be consistent with the propagated prediction with low entropy (\ie, high confidence).}
\label{fig:I-TCR}
\end{figure*} 

\subsection{Intra-domain Regularization}
The intra-domain temporal consistency regularization (I-TCR) aims to minimize the divergence between source and target domains by suppressing the temporal inconsistency across different target frames. As illustrated in Fig.~\ref{fig:I-TCR}, I-TCR guides unconfident target predictions to have similar temporal consistency as confident target predictions. Specifically, it first propagates predictions (of previous frames) forward by using frame-to-frame optical flow estimates, and then forces unconfident predictions in the current frame to be consistent with confident predictions propagated from the previous frame.

In the target domain, we first forward $x^{\mathbb{T}}_{k}$ and $x^{\mathbb{T}}_{k-l}$ to obtain the current prediction $p^{\mathbb{T}}_{k}$, and similarly forward $x^{\mathbb{T}}_{k-1}$ and $x^{\mathbb{T}}_{k-2}$ to obtain the previous prediction $p^{\mathbb{T}}_{k-1}$.  We then adopt FlowNet~\cite{ilg2017flownet} to estimate the optical flow $f_{x^{\mathbb{T}}_{k-1}\rightarrow{}x^{\mathbb{T}}_{k}}$ from $x^{\mathbb{T}}_{k-1}$ to $x^{\mathbb{T}}_{k}$. With the estimated frame-to-frame optical flow $f_{x^{\mathbb{T}}_{k-1}\rightarrow{}x^{\mathbb{T}}_{k}}$, the prediction $p^{\mathbb{T}}_{k-1}$ can be warped to generate the propagated prediction $\hat{p}^{\mathbb{T}}_{k-1}$. 

To force unconfident predictions $p^{\mathbb{T}}_{k}$ in the current frame to be consistent with confident predictions $\hat{p}^{\mathbb{T}}_{k-1}$ propagated from the previous frames, we employ an entropy function $E$ ~\cite{shannon1948mathematical} to estimate the prediction confidence and use the confident prediction $\hat{p}^{\mathbb{T}}_{k-1}$ (\ie, with low entropy) to optimize unconfident prediction $p^{\mathbb{T}}_{k}$ (\ie, with high entropy). Given $p^{\mathbb{T}}_{k}$ and $\hat{p}^{\mathbb{T}}_{k-1}$ from target video frames $X^{\mathbb{T}}$, the I-TCR loss $\mathcal{L}_{itcr}$ can be formulated as follows:
\begin{equation}
\begin{split}
\mathcal{L}_{itcr}(G) = S(E(p^{\mathbb{T}}_{k})-E(\hat{p}^{\mathbb{T}}_{k-1})) |p^{\mathbb{T}}_{k}-\hat{p}^{\mathbb{T}}_{k-1}| .
\end{split}
\label{eq:itcr}
\end{equation}
where $S$ is a signum function which returns 1 if the input is positive or 0 otherwise.

DA-VSN jointly optimizes the source-domain supervised learning (\ie, SSL) and the target-domain unsupervised learning (\ie, C-TCR and I-TCR) as follows:
\begin{equation}
\begin{split}
\min_{G} \max_{D_{st},D_{s}} \mathcal{L}_{ssl} (G) 
& + \lambda_{u} \mathcal{L}_{ctcr} (G,D_{st},D_{s}) \\
& + \lambda_{u} \mathcal{L}_{itcr}(G),
\end{split}
\label{eq:davsn}
\end{equation}
where $\lambda_{u}$ is the weight for balancing the supervised and unsupervised learning in source and target domains.

\section{Experiments}

\subsection{Experimental Setup}
\textbf{Datasets:} Our experiments involve two challenging synthetic-to-real domain adaptive video semantic segmentation tasks: VIPER~\cite{richter2017playing}~$\rightarrow$~Cityscapes-Seq~\cite{cordts2016cityscapes} and SYNTHIA-Seq~\cite{ros2016synthia}~$\rightarrow$~Cityscapes-Seq.
\textbf{Cityscapes-Seq} is a standard benchmark for supervised video semantic segmentation and we use it as the target-domain dataset. It contains $2,975$ and $500$ video sequences for training and evaluation, where each sequence consists of $30$ realistic frames with one ground-truth label provided for the $20^{th}$ frame.
\textbf{VIPER} is used as one source-domain dataset, which contains $133,670$ synthesized video frames with segmentation labels produced by game engines.
\textbf{SYNTHIA-Seq} is used as the other source-domain dataset, which contains $8,000$ synthesized video frames with automatically generated segmentation annotations. The frame resolution is $1024\times2048$, $1080\times1920$ and $760\times1280$ in Cityscapes-Seq, VIPER and SYNTHIA-Seq, respectively.

\textbf{Implementation Details:} We adopt ACCEL~\cite{jain2019accel} as the video semantic segmentation architecture. It consists of two segmentation branches, an optical flow network and a score fusion layer. Each segmentation branch generates single-frame prediction using Deeplab network~\cite{chen2017deeplab} whose backbone is ResNet-101~\cite{he2016deep} pre-trained on ImageNet~\cite{deng2009imagenet}. The optical flow network propagates prediction in the previous frame via FlowNet~\cite{ilg2017flownet} and the score fusion layer adaptively integrates predictions in previous and current frames using a $1\times1$ convolutional layer. 
All the discriminators in our experiments are designed as in DCGAN~\cite{radford2015unsupervised}.
For the efficiency of training and inference, we apply bicubic interpolation to resize every video frame in Cityscapes-Seq and VIPER to $512\times1024$ and $720\times1280$, respectively. Our experiments are built on PyTorch~\cite{paszke2017automatic} and the size of memory usage is below 12 GB. All the models are trained using SGD optimizer with a momentum of $0.9$ and a weight decay of $10^{-4}$. The learning rate is set at $10^{-4}$ and has a polynomial decay with a power of $0.9$. The balancing weights $\lambda_{sa}$, $\lambda_{wd}$, and $\lambda_{u}$ are set as 1, 1 and 0.001, respectively. The mean intersection-over-union (mIoU), the standard evaluation metric in semantic segmentation, is adopted to evaluate all methods.

\renewcommand\arraystretch{1.1}
\begin{table}[!t]
\centering
\begin{small}
\begin{tabular}{p{2cm}|*{4}{p{0.6cm}}|p{0.6cm}}
\toprule
 \multicolumn{6}{c}{\textbf{VIPER~$\rightarrow$~Cityscapes-Seq}} \\
 \midrule
Method &$\mathcal{L}_{ssl}$ &$\mathcal{L}_{sa}$ &$\mathcal{L}_{sta}$ &$\mathcal{L}_{wd}$ &mIoU \\
\midrule
Source only &\checkmark &  &  &  &37.1 \\
SA &\checkmark &\checkmark &  &  &41.6 \\
STA &\checkmark & &\checkmark  &  &43.7 \\
JT &\checkmark &\checkmark &\checkmark  &  &44.2  \\
\textbf{C-TCR} &\checkmark &\checkmark  &\checkmark  &\checkmark  &\textbf{46.5}  \\
\bottomrule
\end{tabular}
\end{small}
\vspace{2pt}
\caption{Ablation study of C-TCR over domain adaptive segmentation task VIPER~$\rightarrow$~Cityscapes-Seq: spatial alignment (SA) and spatial-temporal alignment (STA) both outperform `Source only' greatly. Simple joint training (JT) with STA and SA yields marginal gains over STA, showing that additional spatial alignment does not help much. C-TCR outperforms JT clearly by introducing weight discrepancy loss $L_{wd}$ which forces STA to be independent from SA and focuses more on temporal alignment.}
\label{tab:abla1}
\end{table}

\subsection{Ablation Studies}
We conduct comprehensive ablation studies to examine the effectiveness of our designs and Tables~\ref{tab:abla1}~and~\ref{tab:abla2} show experimental results. Specifically, we trained seven models over the task VIPER~$\rightarrow$~Cityscapes-Seq including: \textbf{1)} Source only that is trained with source data only by using supervised learning loss $\mathcal{L}_{ssl}$; \textbf{2)} ST that performs spatial alignment using adversarial loss $\mathcal{L}_{sa}$ and $\mathcal{L}_{ssl}$; \textbf{3)} STA that performs spatial-temporal alignment using adversarial loss $\mathcal{L}_{sta}$ and $\mathcal{L}_{ssl}$; \textbf{4)} JT that jointly trains SA and STA using $\mathcal{L}_{sa}$, $\mathcal{L}_{sta}$ and $\mathcal{L}_{ssl}$; \textbf{5)} C-TCR that forces STA to focus on cross-domain temporal alignment by introducing the weight discrepancy loss $\mathcal{L}_{wd}$ into JT; \textbf{6)} I-TCR that performs intra-domain adaptation using intra-domain temporal consistency regularization loss $\mathcal{L}_{itcr}$ and $\mathcal{L}_{ssl}$; and \textbf{7)} DA-VSN that integrates C-TCR and I-TCR by using  $\mathcal{L}_{ctcr}$, $\mathcal{L}_{itcr}$ and $\mathcal{L}_{ssl}$.

As shown in Table~\ref{tab:abla1}, both spatial alignment (SA) and spatial-temporal alignment (STA) outperform `Source only' consistently, which verifies the effectiveness of the alignment in spatial and temporal spaces. Specifically, the performance gain of STA is larger than SA, which validates that temporal alignment is important in domain adaptive video segmentation by guiding the target predictions to have similar temporal consistency of source predictions. Joint training (JT) of STA and SA outperforms STA with a marginal performance gain, largely because the spatial-temporal alignment captures spatial alignment already. Cross-domain temporal consistency regularization (C-TCR) improves JT clearly by introducing weight discrepancy loss $L_{wd}$ between discriminators in STA and SA which forces STA to focus on alignment in the temporal space. It also validates the significance of temporal alignment in domain adaptive video semantic segmentation. Similar to C-TCR, intra-domain TCR (I-TCR) outperforms `Source only' with a large margin as shown in Table~\ref{tab:abla2}. This shows the importance of intra-domain adaptation that suppresses temporal inconsistency across target-domain frames. Lastly, DA-VSN produces the best video segmentation, which demonstrates that C-TCR and I-TCR complement with each other.

\renewcommand\arraystretch{1.1}
\begin{table}[!t]
\centering
\begin{small}
\begin{tabular}{p{2cm}|*{3}{p{0.8cm}}|p{0.8cm}}
\toprule
 \multicolumn{5}{c}{\textbf{VIPER~$\rightarrow$~Cityscapes-Seq}} \\
 \midrule
Method &$\mathcal{L}_{ssl}$ &$\mathcal{L}_{ctcr}$ &$\mathcal{L}_{itcr}$ &mIoU \\
\midrule
Source only &\checkmark &  &  &37.1 \\
C-TCR &\checkmark &\checkmark &  &46.5 \\
I-TCR &\checkmark & &\checkmark  &45.9  \\ 
\textbf{DA-VSN} &\checkmark &\checkmark  &\checkmark  &\textbf{47.8}  \\
\bottomrule
\end{tabular}
\end{small}
\vspace{2pt}
\caption{Ablation study of DA-VSN over domain adaptive segmentation task VIPER~$\rightarrow$~Cityscapes-Seq: Cross-domain TCR (C-TCR) and intra-domain TCR (I-TCR) both outperform `Source only' by large margins. In addition, the combination of C-TCR and I-TCR in DA-VSN outperforms either C-TCR or I-TCR clearly, demonstrating the synergic relation of the two designs.}
\label{tab:abla2}
\end{table}

\renewcommand\arraystretch{1.1}
\begin{table*}[!ht]
\centering
\begin{small}
\begin{tabular}{p{2.5cm}|*{15}{p{0.45cm}}p{0.6cm}}
 \toprule
 \multicolumn{17}{c}{\textbf{VIPER~$\rightarrow$~Cityscapes-Seq}} \\
 \midrule
 Methods  &{road} &{side.} &{buil.} &{fence} &{light} &{sign} &{vege.} &{terr.} &{sky} &{pers.} &{car} &{truck} &{bus} &{mot.} &{bike} &mIoU \\
 \midrule
 Source only &56.7 &18.7 &78.7 &6.0 &22.0 &15.6 &81.6 &18.3 &80.4 &59.9 &66.3 &4.5 &16.8 &20.4 &10.3 &37.1 \\
 AdvEnt~\cite{vu2019advent} &78.5 &31.0 &81.5 &22.1 &29.2 &26.6 &81.8 &13.7 &80.5 &58.3 &64.0 &6.9 &38.4 &4.6 &1.3 &41.2 \\
 CBST~\cite{zou2018unsupervised} &48.1 &20.2 &\textbf{84.8} &12.0 &20.6 &19.2 &83.8 &18.4 &\textbf{84.9} &59.2 &71.5 &3.2 &38.0 &23.8 &37.7 &41.7 \\
 IDA~\cite{pan2020unsupervised} &78.7 &33.9 &82.3 &22.7 &28.5 &26.7 &82.5 &15.6 &79.7 &58.1 &64.2 &6.4 &41.2 &6.2 &3.1 &42.0 \\
 CRST~\cite{zou2019confidence} &56.0 &23.1 &82.1 &11.6 &18.7 &17.2 &\textbf{85.5} &17.5 &82.3 &\textbf{60.8} &73.6 &3.6 &38.9 &\textbf{30.5} &\textbf{35.0} &42.4 \\
 FDA~\cite{yang2020fda} &70.3 &27.7 &81.3 &17.6 &25.8 &20.0 &83.7 &\textbf{31.3} &82.9 &57.1 &72.2 &\textbf{22.4} &\textbf{49.0} &17.2 &7.5 &44.4 \\
 \textbf{DA-VSN (Ours)} &\textbf{86.8} &\textbf{36.7} &83.5 &\textbf{22.9} &\textbf{30.2} &\textbf{27.7} &83.6 &26.7 &80.3 &60.0 &\textbf{79.1} &20.3 &47.2 &21.2 &11.4 &\textbf{47.8}\\
\bottomrule
\end{tabular}
\end{small}
\vspace{2pt}
\caption{Quantitative comparisons of DA-VSN with multiple baselines over domain adaptive video segmentation task VIPER~$\rightarrow$~Cityscapes-Seq: DA-VSN outperforms all domain adaptation baselines consistently by large margins.}
\label{tab:bench1}
\end{table*}

\renewcommand\arraystretch{1.1}
\begin{table*}[!ht]
\centering
\begin{small}
\begin{tabular}{p{2.5cm}|*{11}{p{0.7cm}}p{0.8cm}}
 \toprule
 \multicolumn{13}{c}{\textbf{SYNTHIA-Seq~$\rightarrow$~Cityscapes-Seq}} \\
 \midrule
 Methods  &{road} &{side.} &{buil.} &{pole} &{light} &{sign} &{vege.} &{sky} &{pers.} &{rider} &{car} &mIoU \\
 \midrule
 Source only &56.3 &26.6 &75.6 &25.5 &5.7 &15.6 &71.0 &58.5 &41.7 &17.1 &27.9 &38.3 \\
 AdvEnt~\cite{vu2019advent}  &85.7 &21.3 &70.9 &21.8 &4.8 &15.3 &59.5 &62.4 &46.8 &16.3 &64.6 &42.7 \\
 CBST~\cite{zou2018unsupervised} &64.1 &30.5 &78.2 &\textbf{28.9} &\textbf{14.3} &\textbf{21.3} &75.8 &62.6 &46.9 &\textbf{20.2} &33.9 &43.3 \\
 IDA~\cite{pan2020unsupervised} &87.0 &23.2 &71.3 &22.1 &4.1 &14.9 &58.8 &67.5 &45.2 &17.0 &73.4 &44.0 \\
 CRST~\cite{zou2019confidence} &70.4 &31.4 &\textbf{79.1} &27.6 &11.5 &20.7 &\textbf{78.0} &67.2 &\textbf{49.5} &17.1 &39.6 &44.7 \\
 FDA~\cite{yang2020fda} &84.1 &\textbf{32.8} &67.6 &28.1 &5.5 &20.3 &61.1 &64.8 &43.1 &19.0 &70.6 &45.2 \\
 \textbf{DA-VSN (Ours)} &\textbf{89.4} &31.0 &77.4 &26.1 &9.1 &20.4 &75.4 &\textbf{74.6} &42.9 &16.1 &\textbf{82.4} &\textbf{49.5} \\
\bottomrule
\end{tabular}
\end{small}
\vspace{2pt}
\caption{Quantitative comparisons of DA-VSN with multiple baselines over domain adaptive video segmentation task SYNTHIA-Seq~$\rightarrow$~Cityscapes-Seq: DA-VSN outperforms all domain adaptation baselines consistently by large margins.}
\label{tab:bench2}
\end{table*}

\begin{figure*}[!ht]
\centering
\begin{minipage}[h]{0.245\linewidth}
\centering\small {Consecutive video frames}
\end{minipage}
\begin{minipage}[h]{0.245\linewidth}
\centering\small {Ground Truth$^{\ast}$} 
\end{minipage}
\begin{minipage}[h]{0.245\linewidth}
\centering\small {FDA~\cite{yang2020fda}}
\end{minipage}
\begin{minipage}[h]{0.245\linewidth}
\centering\small {\textbf{DA-VSN (Ours)}} 
\end{minipage}
\vspace{2pt}
\centering
\begin{minipage}[h]{0.245\linewidth}
\centering\includegraphics[width=.99\linewidth]{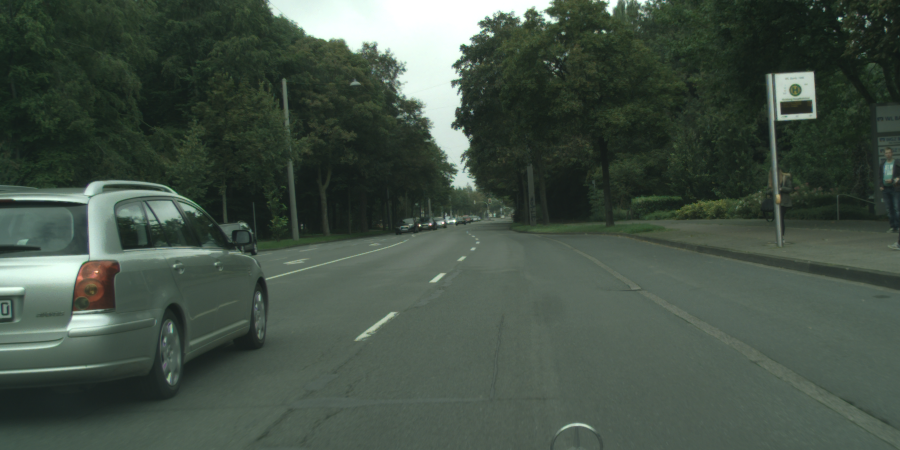}
\end{minipage}
\begin{minipage}[h]{0.245\linewidth}
\centering\includegraphics[width=.99\linewidth]{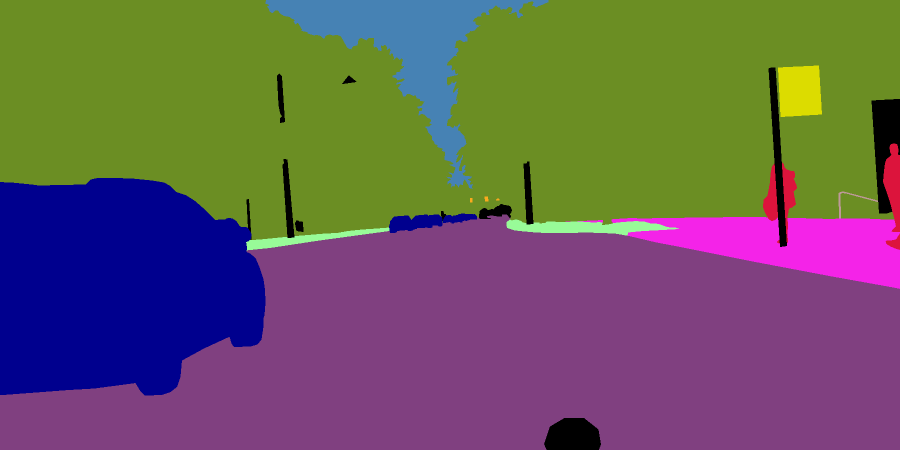}
\end{minipage}
\begin{minipage}[h]{0.245\linewidth}
\centering\includegraphics[width=.99\linewidth]{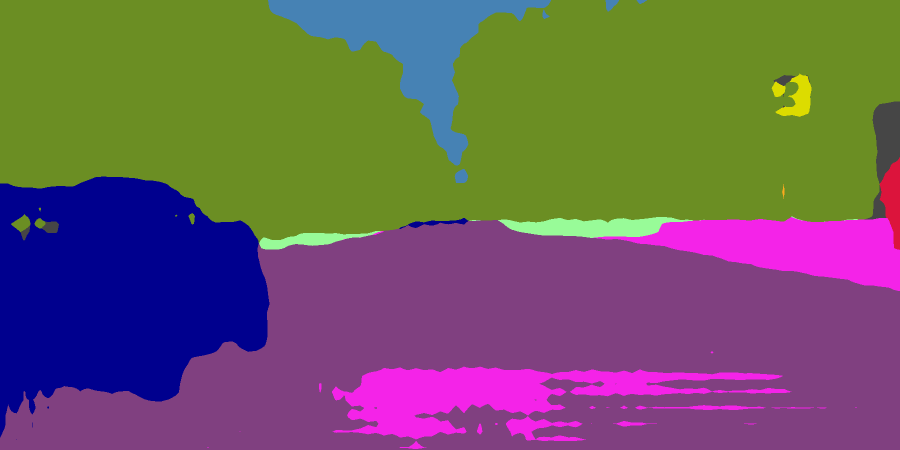}
\end{minipage}
\begin{minipage}[h]{0.245\linewidth}
\centering\includegraphics[width=.99\linewidth]{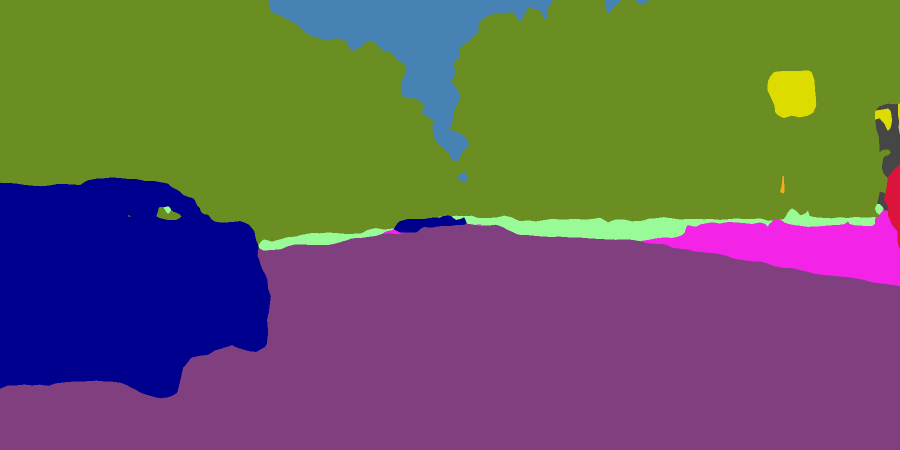}
\end{minipage}
\centering
\vspace{2pt}
\centering
\begin{minipage}[h]{0.245\linewidth}
\centering\includegraphics[width=.99\linewidth]{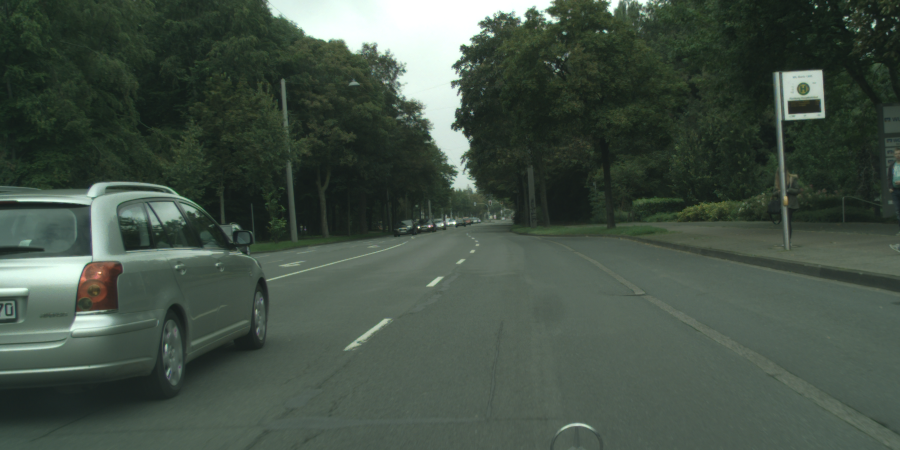}
\end{minipage}
\begin{minipage}[h]{0.245\linewidth}
\centering\includegraphics[width=.99\linewidth]{figures/fig_results/munster_000001_000019_gt.png}
\end{minipage}
\begin{minipage}[h]{0.245\linewidth}
\centering\includegraphics[width=.99\linewidth]{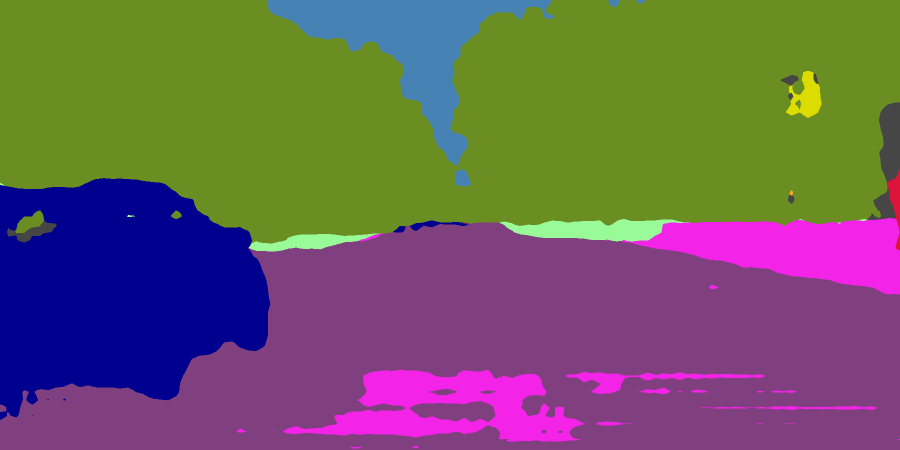}
\end{minipage}
\begin{minipage}[h]{0.245\linewidth}
\centering\includegraphics[width=.99\linewidth]{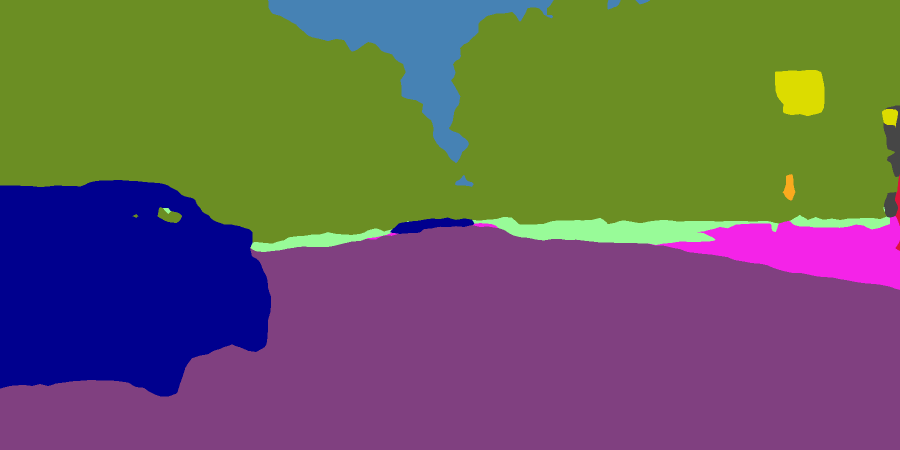}
\end{minipage}
\vspace{2pt}
\centering
\begin{minipage}[h]{0.245\linewidth}
\centering\includegraphics[width=.99\linewidth]{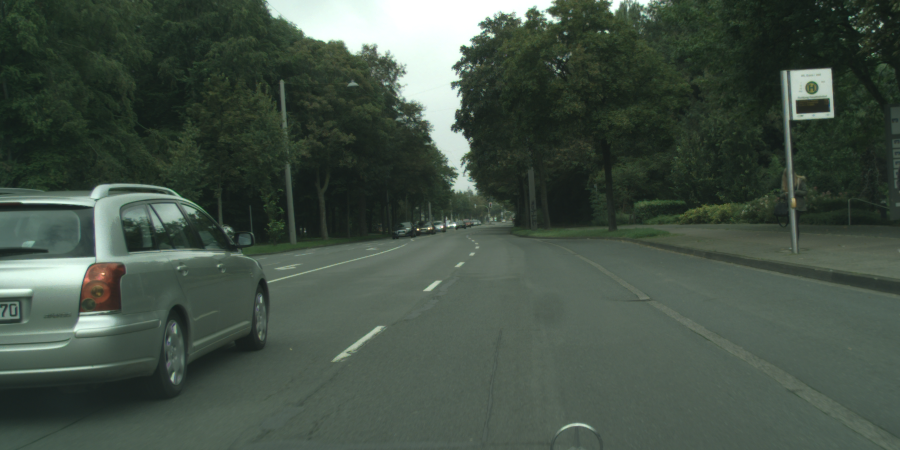}
\end{minipage}
\begin{minipage}[h]{0.245\linewidth}
\centering\includegraphics[width=.99\linewidth]{figures/fig_results/munster_000001_000019_gt.png}
\end{minipage}
\begin{minipage}[h]{0.245\linewidth}
\centering\includegraphics[width=.99\linewidth]{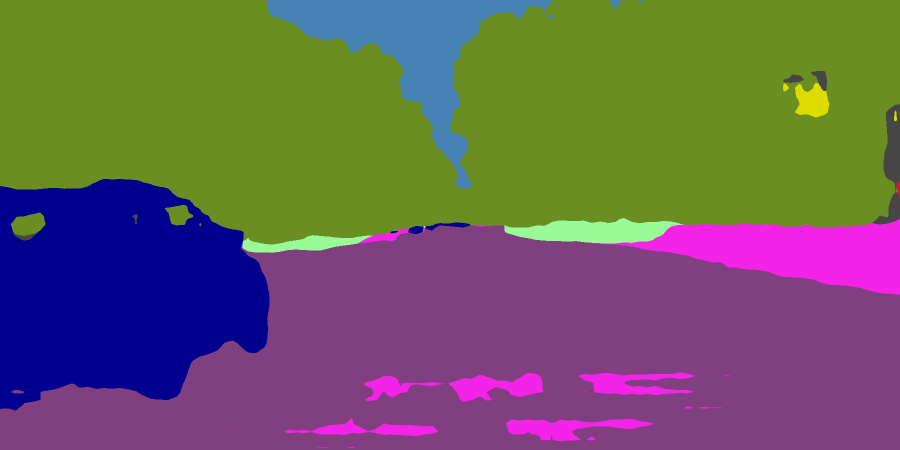}
\end{minipage}
\begin{minipage}[h]{0.245\linewidth}
\centering\includegraphics[width=.99\linewidth]{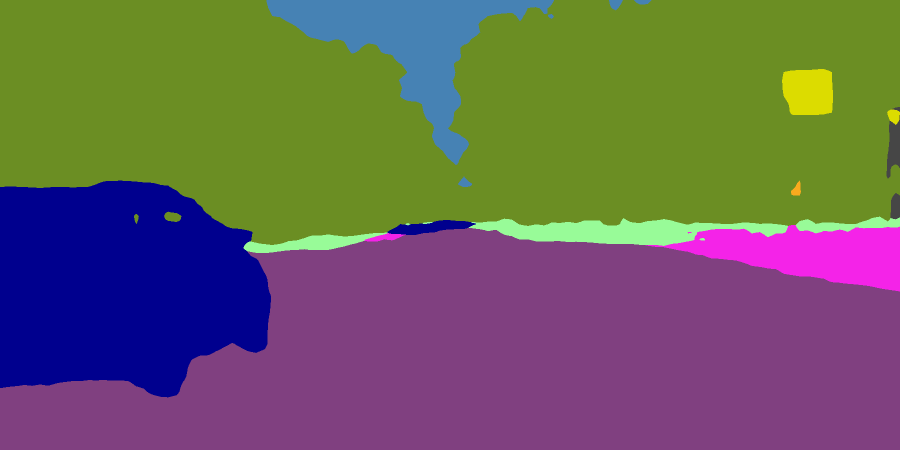}
\end{minipage}
\centering
\caption{Qualitative comparison of DA-VSN with the best-performing baseline \textit{FDA}~\cite{yang2020fda} over domain adaptive video segmentation task \enquote{VIPER~$\rightarrow$~Cityscapes-Seq}: DA-VSN produces more accurate pixel-wise segmentation predictions with higher temporal consistency across consecutive video frames as shown in rows 1-3. 
Since Cityscapes-Seq only provides ground-truth label of one frame per $30$ consecutive frames, we show the same ground truth for all rows. The \textit{Ground Truth$^{\ast}$} denotes the ground truth annotated for the video frame in Row 2.
Best viewed in color.}
\label{fig:result}
\end{figure*}

\subsection{Comparison with Baselines}
\label{Comparison with Baselines}
Since few works study domain adaptive video semantic segmentation, we quantitatively compare DA-VSN with multiple domain adaptation baselines~\cite{vu2019advent, zou2018unsupervised, pan2020unsupervised,zou2019confidence,yang2020fda} that achieved superior performance in domain adaptive image segmentation. We apply these approaches to the domain adaptive video segmentation task by simply replacing their image segmentation model by video segmentation model and performing domain alignment as in~\cite{vu2019advent, zou2018unsupervised, pan2020unsupervised, zou2019confidence,yang2020fda}. The comparisons are performed over two synthetic-to-real domain adaptive video segmentation tasks as shown in Tables~\ref{tab:bench1} and \ref{tab:bench2}. As the two tables show, the proposed method outperforms all the domain adaptation baselines consistently with large margins. 

We also perform qualitative comparisons over the video segmentation task  VIPER~$\rightarrow$~Cityscapes-Seq. We compare the proposed DA-VSN with the best-performing baseline FDA~\cite{yang2020fda} as illustrated in Fig.~\ref{fig:result}. We can see that the qualitative results are consistent with the quantitative results in Table~\ref{tab:bench1}. Specifically, our method can generate better segmentation results with higher temporal consistency across consecutive video frames. The excellent segmentation performance is largely attributed to the proposed temporal consistency regularization which minimizes the divergence of temporal consistency across different domains and different target-domain video frames.

\begin{figure*}[!ht]
\centering
\begin{minipage}[c]{0.32\linewidth}
\centering\small {Source only}
\end{minipage}
\begin{minipage}[c]{0.32\linewidth}
\centering\small {FDA~\cite{yang2020fda}} 
\end{minipage}
\begin{minipage}[c]{0.32\linewidth}
\centering\small {\textbf{DA-VSN (Ours)}}
\end{minipage}
\begin{minipage}[h]{0.32\linewidth}
\centering\includegraphics[width=.99\linewidth]{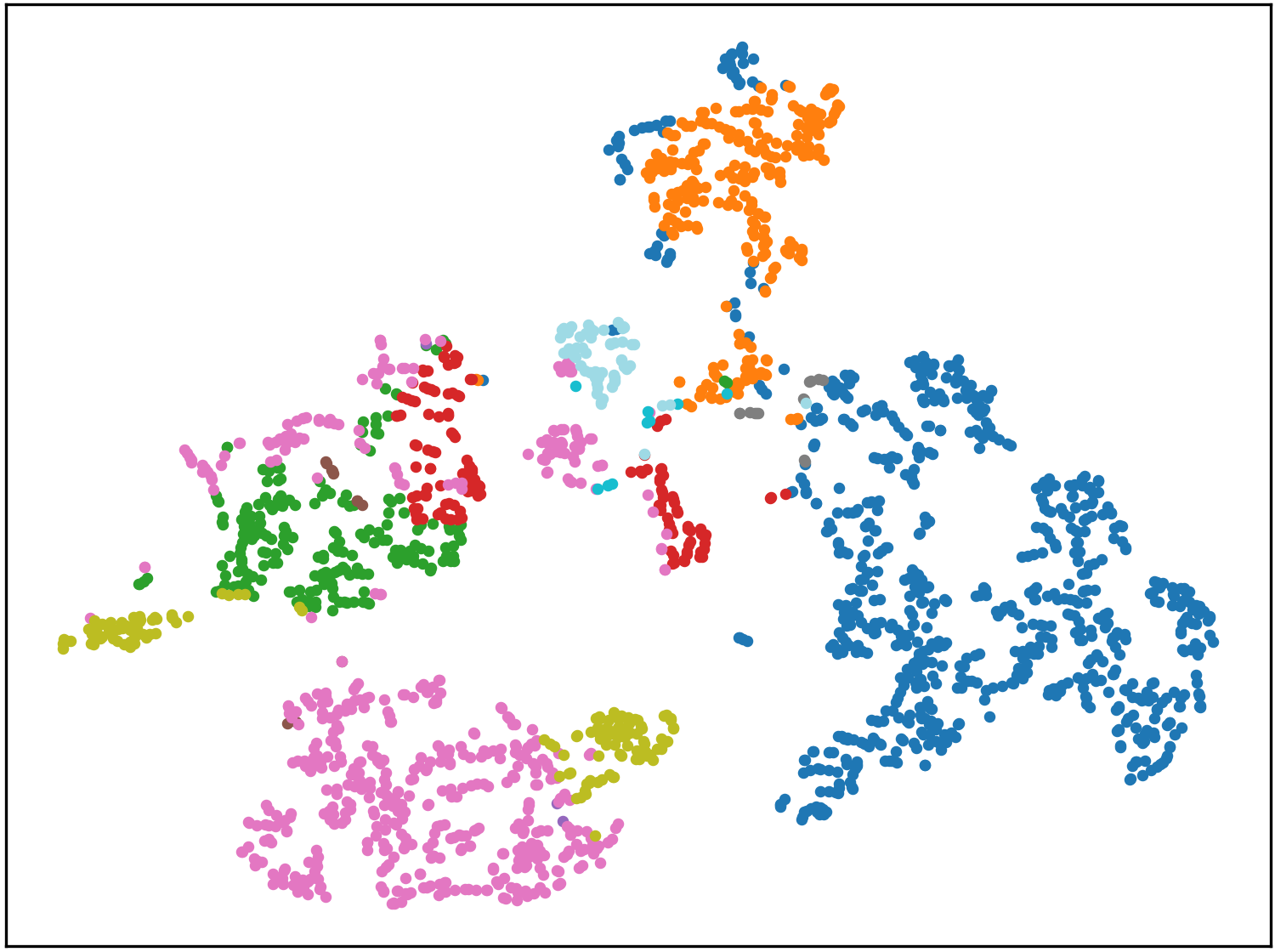}
\end{minipage}
\begin{minipage}[h]{0.32\linewidth}
\centering\includegraphics[width=.99\linewidth]{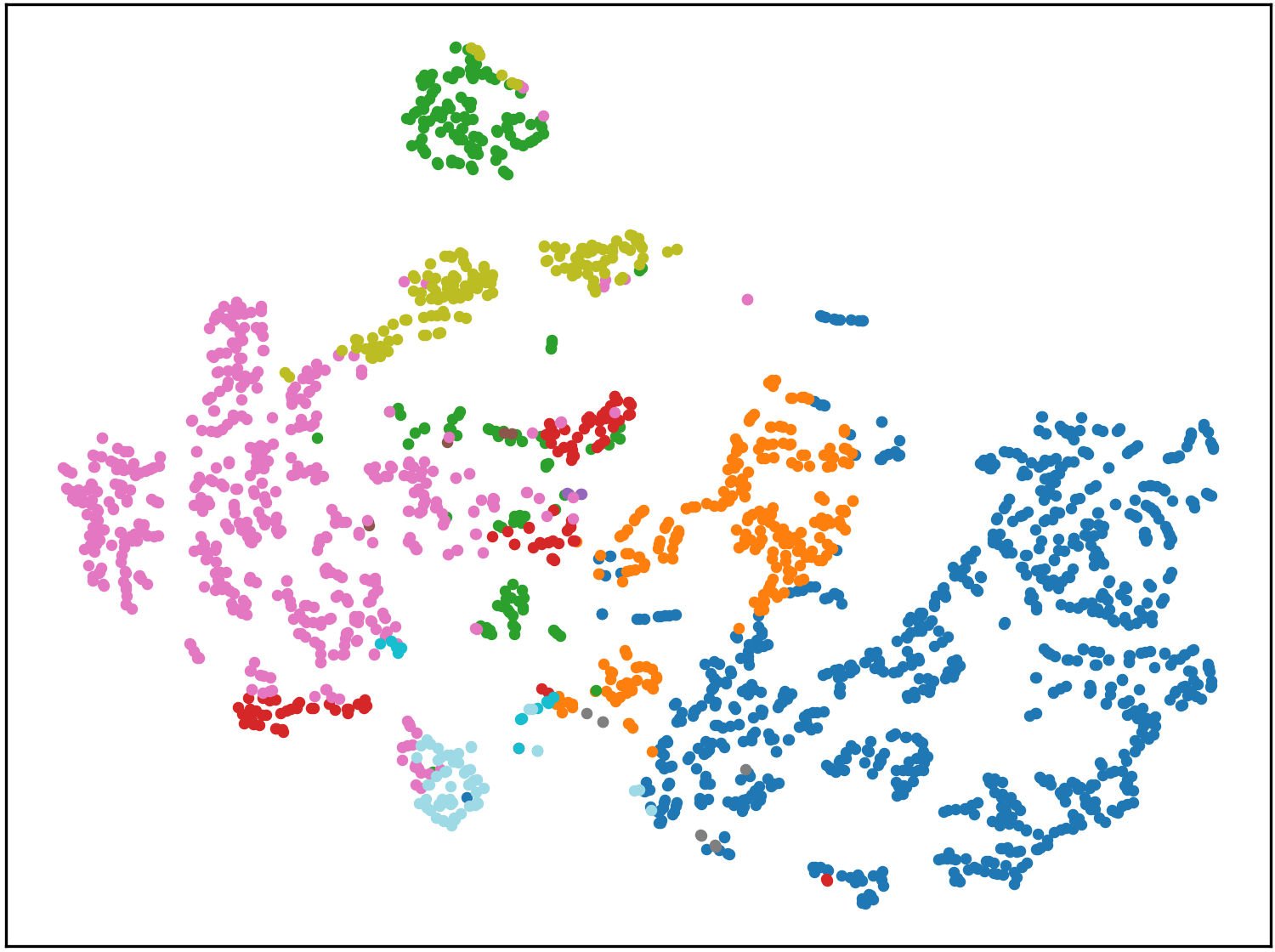}
\end{minipage}
\begin{minipage}[h]{0.32\linewidth}
\centering\includegraphics[width=.99\linewidth]{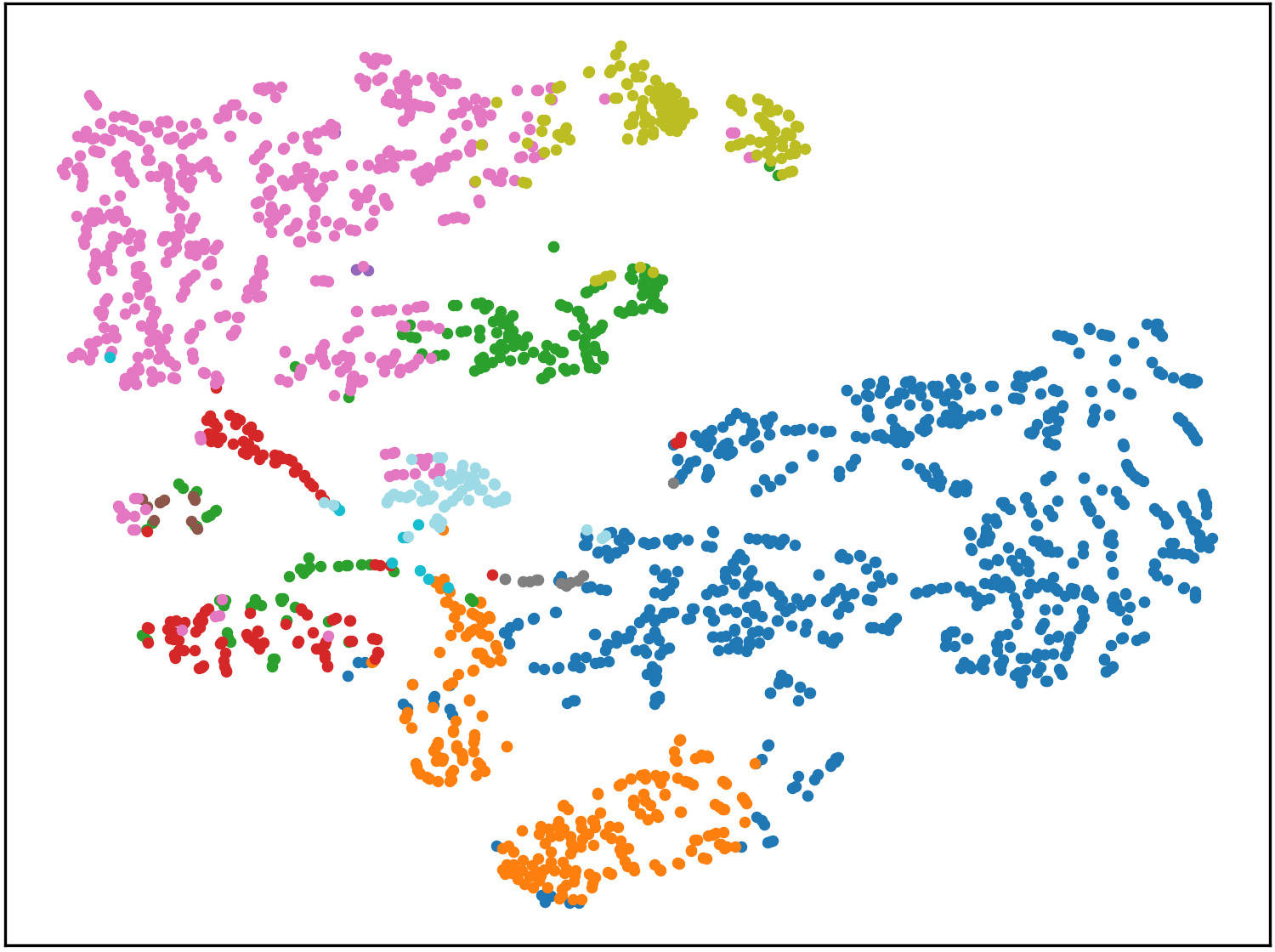}
\end{minipage}
\centering
\begin{minipage}[h]{0.32\linewidth}
\centering\small {$\sigma_{\text{inter}}^{2}=761.3$, $\sigma_{\text{intra}}^{2}=331.7$}
\end{minipage}
\begin{minipage}[h]{0.32\linewidth}
\centering\small {$\sigma_{\text{inter}}^{2}=781.2$, $\sigma_{\text{intra}}^{2}=239.0$} 
\end{minipage}
\begin{minipage}[h]{0.32\linewidth}
\centering\small {$\sigma_{\text{inter}}^{2}=854.26$, $\sigma_{\text{intra}}^{2}=186.2$}
\end{minipage}
\caption{Visualization the distribution of temporal feature representations in the target domain via t-SNE~\cite{maaten2008visualizing}: We calculate inter-class variance $\sigma_{\text{inter}}^{2}$ and intra-class variance $\sigma_{\text{intra}}^{2}$ of temporal features, \ie, the stacked feature maps from two consecutive frames. It can be observed that the proposed DA-VSN outperforms `Source only’ model and FDA~\cite{yang2020fda} clearly. Evaluation is conducted on the domain adaptive video segmentation task \enquote{VIPER~$\rightarrow$~Cityscapes-Seq}. Note that different colors denote different classes and best viewed in color.}
\label{fig:tsne}
\end{figure*}

\subsection{Discussion}

\textbf{Feature Visualization:}
In the Section~\ref{Comparison with Baselines}, we have demonstrated that the proposed DA-VSN has achieved superior performance in domain adaptive video segmentation as compared with multiple baselines. To further study the properties of DA-VSN, we use t-SNE~\cite{maaten2008visualizing} to visualize the distribution of target-domain temporal feature representations from different domain adaptive video segmentation methods, where the inter-class and intra-class variances are computed for quantitative analysis. As shown in Fig.~\ref{fig:tsne}, DA-VSN produces the most discriminative target-domain temporal features with the largest inter-class variance and the smallest intra-class variance, as compared with `Source only' and FDA~\cite{yang2020fda}. 

\renewcommand\arraystretch{1.1}
\begin{table}[!t]
\centering
\begin{small}
\begin{tabular}{p{2cm}|*{3}{p{1.2cm}}}
\toprule
 \multicolumn{4}{c}{\textbf{VIPER~$\rightarrow$~Cityscapes-Seq}} \\
 \midrule
\multicolumn{1}{c|}{Method} &\multicolumn{1}{c}{Base}  &\multicolumn{1}{c}{+ DA-VSN}
&\multicolumn{1}{c}{Gain} \\
\midrule
FDA~\cite{yang2020fda} &\multicolumn{1}{c}{44.4} &\multicolumn{1}{c}{48.5} &\multicolumn{1}{c}{+4.1} \\
IDA~\cite{pan2020unsupervised} &\multicolumn{1}{c}{42.0} &\multicolumn{1}{c}{49.9} &\multicolumn{1}{c}{+7.9} \\
CBST~\cite{zou2018unsupervised} &\multicolumn{1}{c}{41.7} &\multicolumn{1}{c}{50.2} &\multicolumn{1}{c}{+8.5} \\
CRST~\cite{zou2019confidence} &\multicolumn{1}{c}{42.4} &\multicolumn{1}{c}{51.3} &\multicolumn{1}{c}{+8.9} \\
\bottomrule
\end{tabular}
\end{small}
\vspace{2pt}
\caption{The proposed DA-VSN complements with multiple domain adaption baselines over domain adaptive video segmentation task VIPER~$\rightarrow$~Cityscapes-Se: \textit{DA-VSN} can be easily incorporated into state-of-the-art domain adaptive image segmentation methods~\cite{zou2018unsupervised,zou2019confidence,pan2020unsupervised,yang2020fda} with consistent performance improvement.}
\label{tab:comp}
\end{table}

\textbf{Complementary Studies:}
We also investigate whether the proposed DA-VSN can complement with multiple domain adaptation baselines~\cite{zou2018unsupervised,pan2020unsupervised,zou2019confidence,yang2020fda} (as described in Section~\ref{Comparison with Baselines}) over domain adaptive video segmentation task. To conduct this experiment, we integrate our proposed temporal consistency regularization components (DA-VSN) into these baselines and Table \ref{tab:comp} shows the segmentation results of the newly trained models. It can be seen that the incorporation of DA-VSN improves video segmentation performance greatly across all the baselines, which shows that DA-VSN is complementary to the domain adaptation methods that minimize domain discrepancy via image translation (\eg, FDA~\cite{yang2020fda}), adversarial learning (\eg, AdvEnt~\cite{vu2019advent}) and self-training (\eg, CBST~\cite{zou2018unsupervised} and CRST~\cite{zou2019confidence}).

\renewcommand\arraystretch{1.1}
\begin{table}[!t]
\centering
\begin{small}
\begin{tabular}{p{2cm}|*{3}{p{1cm}}}
\toprule
 \multicolumn{4}{c}{\textbf{VIPER~$\rightarrow$~Cityscapes-Seq}} \\
 \midrule
Architectures &\multicolumn{1}{c}{Source only} &\multicolumn{1}{c}{DA-VSN} &\multicolumn{1}{c}{Gain} \\
\midrule
NetWarp~\cite{gadde2017semantic}  &\multicolumn{1}{c}{36.5} &\multicolumn{1}{c}{47.2} &\multicolumn{1}{c}{+10.7} \\
TDNet~\cite{hu2020temporally} &\multicolumn{1}{c}{37.6} &\multicolumn{1}{c}{47.9} &\multicolumn{1}{c}{+10.3} \\
ESVS~\cite{liu2020efficient} &\multicolumn{1}{c}{38.2} &\multicolumn{1}{c}{48.1} &\multicolumn{1}{c}{+9.9} \\
\bottomrule
\end{tabular}
\end{small}
\vspace{2pt}
\caption{DA-VSN can work with different video semantic segmentation architectures: \textit{DA-VSN} can work with different video segmentation architectures (e.g. Netwarp~\cite{gadde2017semantic}, TDNet~\cite{hu2020temporally} and ESVS~\cite{liu2020efficient}) with consistent performance improvement as compared with \textit{Source only} over the domain adaptive video segmentation task \enquote{VIPER~$\rightarrow$~Cityscapes-Seq}.}
\label{tab:diff}
\end{table}

\textbf{Different Video Segmentation Architectures:}
We further study whether DA-VSN can work well with different video semantic segmentation architectures. Three widely adopated video segmentation architectures (\ie, Netwarp~\cite{gadde2017semantic}, TDNet~\cite{hu2020temporally} and ESVS~\cite{liu2020efficient}) are used in this experiments. As shown in Table \ref{tab:diff}, the proposed DA-VSN outperforms the `Source only' consistently with large margins. This experiment shows that our method performs excellently with different video semantic segmentation architectures that exploits temporal relations via feature propagation~\cite{gadde2017semantic}, attention propagation~\cite{hu2020temporally}, and temporal consistency constraint~\cite{liu2020efficient}.

\section{Conclusion}
This paper presents a domain adaptive video segmentation network that introduces cross-domain temporal consistency regularization (TCR) and intra-domain TCR to address domain shift in videos. Specifically, cross-domain TCR performs spatial and temporal alignment that guides the target video predictions to have similar temporal consistency as the source video predictions. Intra-domain TCR directly minimizes the discrepancy of temporal consistency across different target video frames. Extensive experiments demonstrate the superiority of our method in domain adaptive video segmentation. In the future, we will adapt the idea of temporal consistency regularization to other video domain adaptation tasks such as video instance segmentation and video panoptic segmentation.

{\small
\bibliographystyle{ieee_fullname}
\bibliography{manuscript}
}

\end{document}